\def\year{2022}\relax
\documentclass[letterpaper]{article} % DO NOT CHANGE THIS
\usepackage{aaai22}  % DO NOT CHANGE THIS
\usepackage{times}  % DO NOT CHANGE THIS
\usepackage{helvet}  % DO NOT CHANGE THIS
\usepackage{courier}  % DO NOT CHANGE THIS
\usepackage[hyphens]{url}  % DO NOT CHANGE THIS
\usepackage{graphicx} % DO NOT CHANGE THIS
\usepackage{natbib}  % DO NOT CHANGE THIS AND DO NOT ADD ANY OPTIONS TO IT
\usepackage{caption} % DO NOT CHANGE THIS AND DO NOT ADD ANY OPTIONS TO IT
\usepackage{float} \usepackage{graphicx}
\usepackage{booktabs}
\DeclareCaptionStyle{ruled}{labelfont=normalfont,labelsep=colon,strut=off} % DO NOT CHANGE THIS
\usepackage{algorithm}
\usepackage{algorithmic}

\usepackage{newfloat}
\usepackage{listings}
\floatstyle{ruled}
\newfloat{listing}{tb}{lst}{}
\floatname{listing}{Listing}
\usepackage{kotex} 
\usepackage{todonotes}
\newcommand*{\affaddr}[1]{#1} % No op here. Customize it for different styles.
\newcommand*{\affmark}[1][*]{\textsuperscript{#1}}

\usepackage{xcolor, soul}

\title{\textsc{FreeTalky}: Don’t Be Afraid! Conversations Made Easier by a \\Humanoid Robot using Persona-based Dialogue}
\author{%
Chanjun Park\affmark[1]\equalcontrib, Yoonna Jang\affmark[1]\equalcontrib, Seolhwa Lee\affmark[2]\equalcontrib, Sungjin Park\affmark[3]\equalcontrib, and Heuiseok Lim\affmark[1]\thanks{This author is the corresponding author.}\\
\affaddr{\affmark[1]Korea University}\\
\affaddr{\affmark[2]University of Copenhagen}\\
\affaddr{\affmark[3]NAVER Corp.}\\
\{bcj1210, morelychee, limhseok\}@korea.ac.kr, sele@di.ku.dk, sungjin.park@navercorp.com }

\begin{document}

\maketitle

\begin{abstract}
We propose a deep learning-based foreign language learning platform, named \textsc{FreeTalky}, for people who experience anxiety dealing with foreign languages, by employing a humanoid robot NAO and various deep learning models. A persona-based dialogue system that is embedded in NAO provides an interesting and consistent multi-turn dialogue for users. Also, an grammar error correction system promotes improvement in grammar skills of the users. Thus, our system enables personalized learning based on persona dialogue and facilitates grammar learning of a user using grammar error feedback. Furthermore, we verified whether \textsc{FreeTalky} provides practical help in alleviating xenoglossophobia by replacing the real human in the conversation with a NAO robot, through human evaluation.
\end{abstract}

\section{Introduction}
Following multiculturalism and hyper-connectivity, many societies value multilingual skills as a demonstration of the ability to work effectively with customers, clients, and businesses of various countries and cultures~\cite{damari2017demand}. In this context, native-speaking classes and foreign language phone calls for foreign language learning are available in most countries. 

However, the costs are very high and some learners experience a mental block in learning foreign languages ({\em{i.e.,}} foreign language anxiety, also known as xenoglossophobia). It can be described as a feeling of uneasiness, worry, nervousness, and apprehension that is experienced while learning or using a second or foreign language~\cite{horwitz1986foreign, donmus2010use}.

To address these issues, we propose \textsc{FreeTalky}, a software platform that allows users to communicate comfortably as if they were conversing with a native speaker by conversing with a robot instead through realistic free talking. With \textsc{FreeTalky}, users obtain a learning effect beyond the traditional classroom learning methods, by experiencing direct conversations without the mental pressure, rather than by only seeing and hearing. 

The \textsc{FreeTalky} platform combines speech recognition, a persona-based dialogue system, English grammar error correction, and a humanoid robot NAO, for English learning that is user-friendly and practically helpful, by reflecting user perspectives. Specifically, the persona-based dialogue system provides customized multi-turn dialogue, while the English grammar error correction software provides customized grammar correction. These features allow users to have user-friendly conversations with the robot, learn foreign languages effectively, and even gain assistance with their xenoglossophobia. The contributions of this study are as follows.

\begin{itemize}
\item To the best of our knowledge, \textsc{FreeTalky} is the first deep learning-based conversation platform for alleviating xenoglossophobia.

\item The persona-based conversation system in \textsc{FreeTalky} allows users to customize foreign language learning and provides practical feedback on grammatical errors through a deep neural network-based grammar error correction software.

\item Through an evaluation and survey of real users of FreeTalky, the validity and feasibility of FreeTalky were verified, and various experimental empirical analyses and error analyses were performed.
\end{itemize}

\begin{figure*}[t!]
\begin{center}
\includegraphics[width=0.85\linewidth]{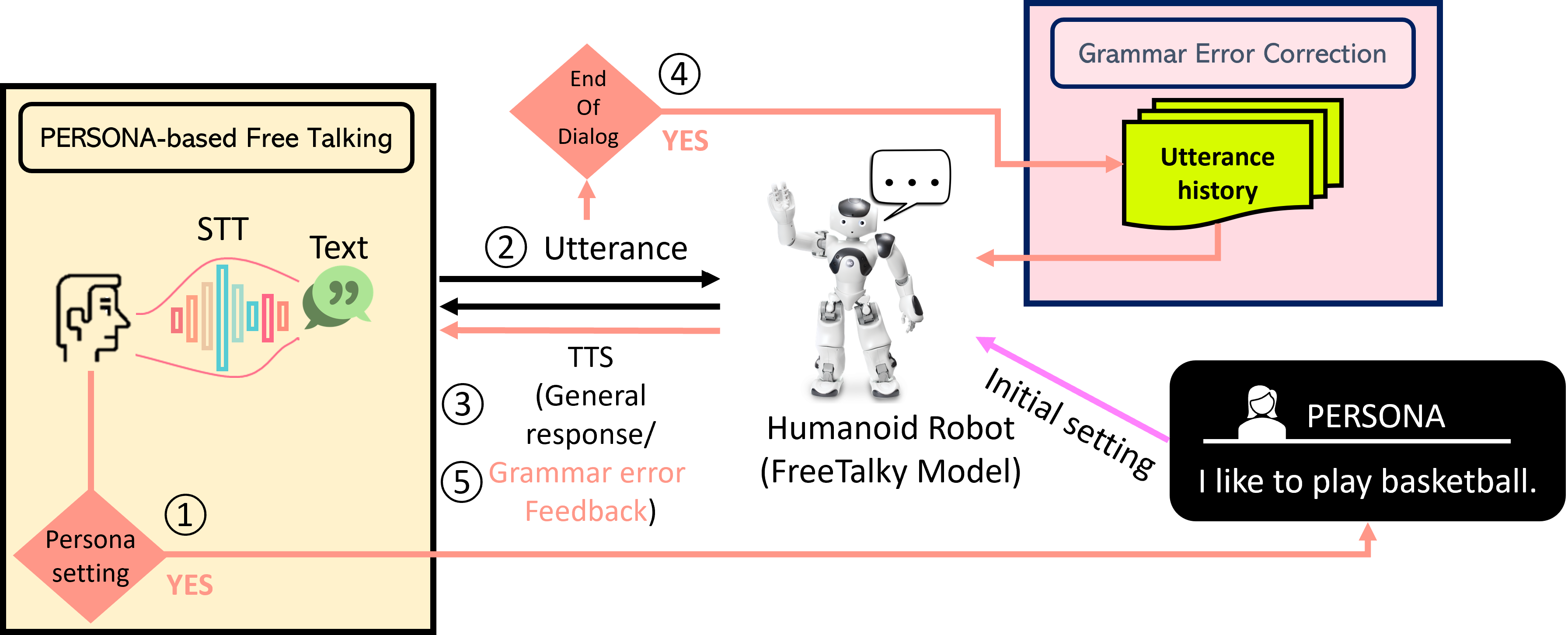}
\end{center}
\caption{Overview of \textsc{FreeTalky} process.} 
\label{fig:freetalky}
\end{figure*}

\section{Related Work}
To the best of our knowledge, there is no foreign language learning software for alleviating xenoglossophobia. In fact, a majority of the commercialized software products for foreign language learning are developed for the general public, with the primary objective of improving English language skills, and do not consider xenoglossophobia.

There are three major foreign language learning platforms: Andy \footnote{\url{https://andychatbot.com/}}, Cake \footnote{\url{https://mycake.me/}}, and SpeakNow \footnote{\url{https://www.speaknow.ai}}. However, despite the commercialized service system, there is considerable room for improvement.

First, Andy does not provide contextual dialogue, and users cannot receive evaluation or feedback on their learning. Second, Cake provides feedback on pronunciation and expressiveness; however, there are no means to have a natural conversation through an actual dialogue system. Additionally, it only provides traditional evaluation methods, such as word quizzes and fill in the blanks. Further, there is no feature that provides practical feedback for English grammar. Finally, SpeakNow, an educational English conversation application using 3D avatars, provides evaluation of grammar accuracy, expressiveness, and pronunciation, exploiting various conversation situations and users' voices. However, it is not a type of free conversational model, as it only provides a service that selects and answers a given sentence, or reads the given sentences.

In light of these limitations, we propose \textsc{FreeTalky}, which is equipped with a persona conversation system that enables personalized learning and an English grammar corrector that provides practical learning feedback.

\section{\textsc{FreeTalky}}
\paragraph{Workflow of \textsc{FreeTalky}} 
In this section, we describe the overall workflow of \textsc{FreeTalky}, as shown in Figure~\ref{fig:freetalky}. The process is explained step by step as follows. 

\begin{enumerate}
\item The user sets the persona of the robot (\textit{e.g.,} (1) [\textit{I like to play basketball., ...}], (2) [\textit{I have two daughters., ...}], (3) [\textit{I am a graduate student., ...}]) by selecting a number. Each number is assigned a randomly chosen list of personas, and this information is entered into the system. Automatic persona setting is provided for the convenience of the user; however, the users can set the persona of the robot themselves as well.

\item The user begins the conversation with the NAO robot to practice speaking in English. The input of the user is further converted into text using the speech-to-text (STT) module, and passed on to the dialogue system. The dialogue model inputs the converted text with persona sentences and previous history of the dialogue. Next, the model generates a response.

\item The generated response is converted into speech using the text-to-speech (TTS) module of the NAO robot, for it to provide the answers to the user. Through this process, the user is able to receive the output from the model in the form of human voice. 

\item When the conversation is completed by the end command ({\em{i.e.,} Bye.}), all utterances are transmitted to the grammar error correction (GEC) module, where the grammatical errors in the utterances are detected.

\item The robot provides feedback on the sentences that include grammatical errors, if there are any. By receiving feedback, the users learn English grammar, in addition to experiencing a seamless conversation in English.
\end{enumerate}

\paragraph{NAO Robot}
According to ~\citet{shamsuddin2011humanoid}, NAO is a humanoid robot that has been purposely designed to seem approachable. We embedded our \textsc{FreeTalky} in the NAO robot to allow the robot to be able to communicate with people more easily and effectively. By utilizing this humanoid robot as an educational assistant, people who are learning the English language would experience more natural conversations, rather than studying alone. 

\paragraph{Speech-to-Text} We adopted an off-the-shelf STT module known as CLOVA speech recognition (CSR)~\cite{ha2020clovacall} API of Naver. It obtains the speech input of the user and converts it into text. Further, it sends the converted text to the dialogue system of the model.

\paragraph{Text-to-Speech} The TTS system is based on ALTextToSpeech, which is a built-in TTS system of the NAO robot. This module is used for providing feedback to the users on their grammatical errors. In addition, the robot alters its emotion and voice, which allows the users to experience a life-like conversation with foreigners.

\begin{figure}[h!]
\begin{center}
\includegraphics[width=0.9\linewidth]{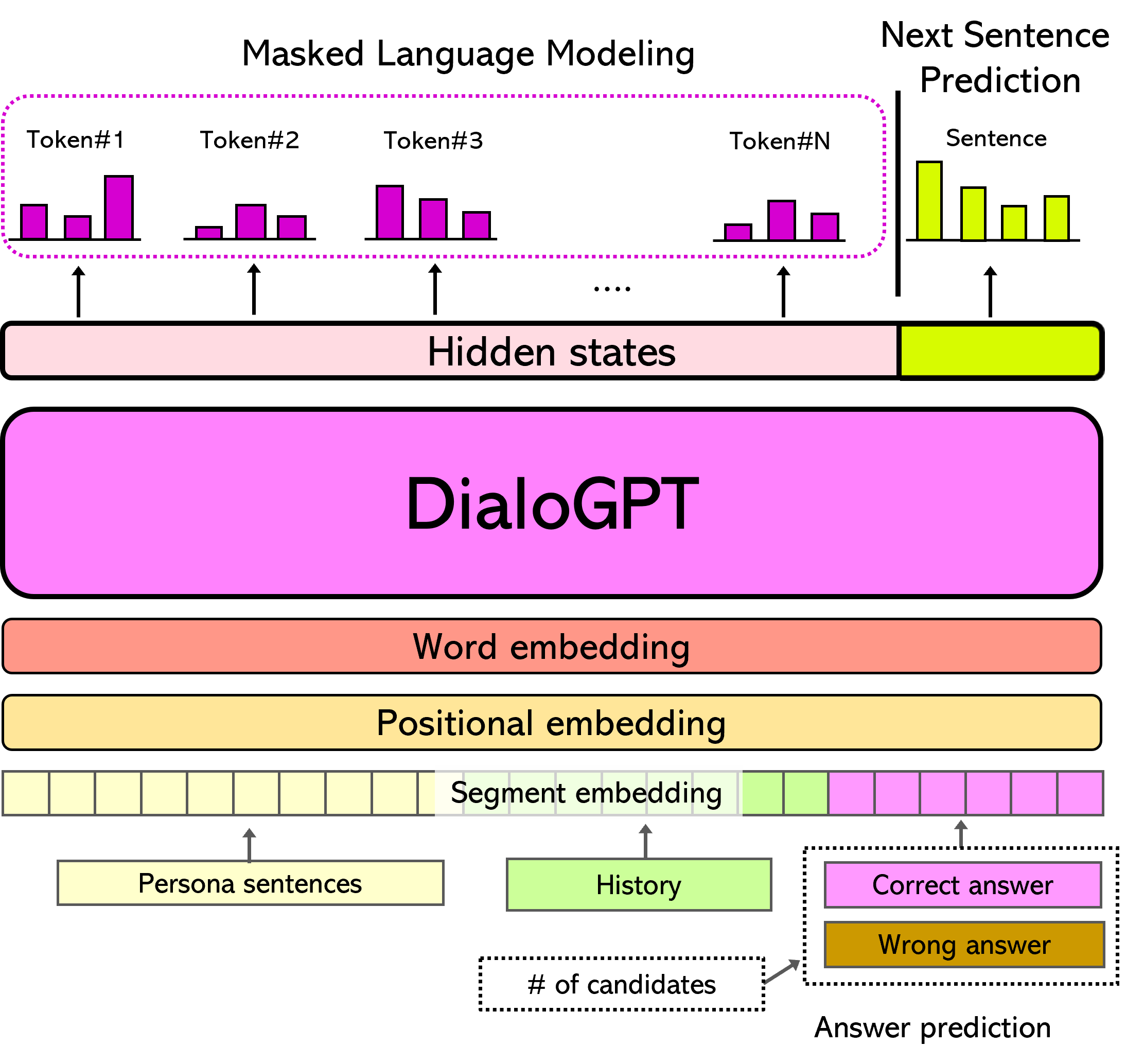}
\end{center}
\caption{Overview of persona-based free talking model in \textsc{FreeTalky}.}
\label{fig:persona_freetalky}
\end{figure}

\paragraph{Persona-based Free Talking Dialogue}
According to Grice's maxims, there are several maxims that need to be followed to ensure a good conversation~\cite{dale1995computational,looi2012applying}. They are maxims of quantity, quality, relation, and manners. 

To reflect these maxims, \textsc{FreeTalky} adopts a persona-based dialogue model. By assigning a persona to the system, it can provide consistent and friendly responses to the users. The users can practice conversing in English with the NAO robot similarly as they would converse with people. Additionally, they can set up the persona of the robot such that it includes their various interests or conversational topics.

Most existing foreign language learning systems provide only a single-turn dialogue. However, we deployed a multi-turn free-talking model using the pre-trained language model~\cite{zhang2019dialogpt} for a more realistic conversation. It was trained on a massive dialogue corpus to allow the model to utilize the learned dialogue patterns and construct colloquial expressions. We adopted PERSONA-CHAT~\cite{zhang2018personalizing} as the training data for transfer learning of the model. 

In the transfer-learning phase, the concatenations of persona sentences, dialogue history, and the user's utterances were used as the input. With this input, the model learned the patterns inherent in dialogue data. As shown in Figure \ref{fig:persona_freetalky}, the model was trained with two objective functions: next sentence prediction and masked language modeling. With the masked language modeling, the model learned next token generation in an auto-regressive manner. Simultaneously, it learned which answer is suitable for the next utterance among 20 candidates, using the next sentence prediction function. 

The evaluation metrics for the model are hits@1 and perplexity. Hits@1 is used to measure the accuracy of the correct answer being chosen among 20 candidate answers. Perplexity is a metric for evaluating a language model for its ability to predict the next token. Low perplexity indicates low congestion and high performance of language models. In the results obtained from the qualitative analysis, hits@1 was 78.68 and perplexity was 14.19, when evaluated on the validation set of the PERSONA-CHAT dataset \cite{zhang2018personalizing}. In addition, to handle ethical issues (e.g. hate speech, offensive words) which frequently arises from the language models due to the biases in toxic language datasets \cite{sap2019risk}, we regenerate the sequence, if the generated sequence has certain words in our pre-built list of offensive words. 

\begin{figure}[h!]
\begin{center}
\includegraphics[width=0.9\linewidth]{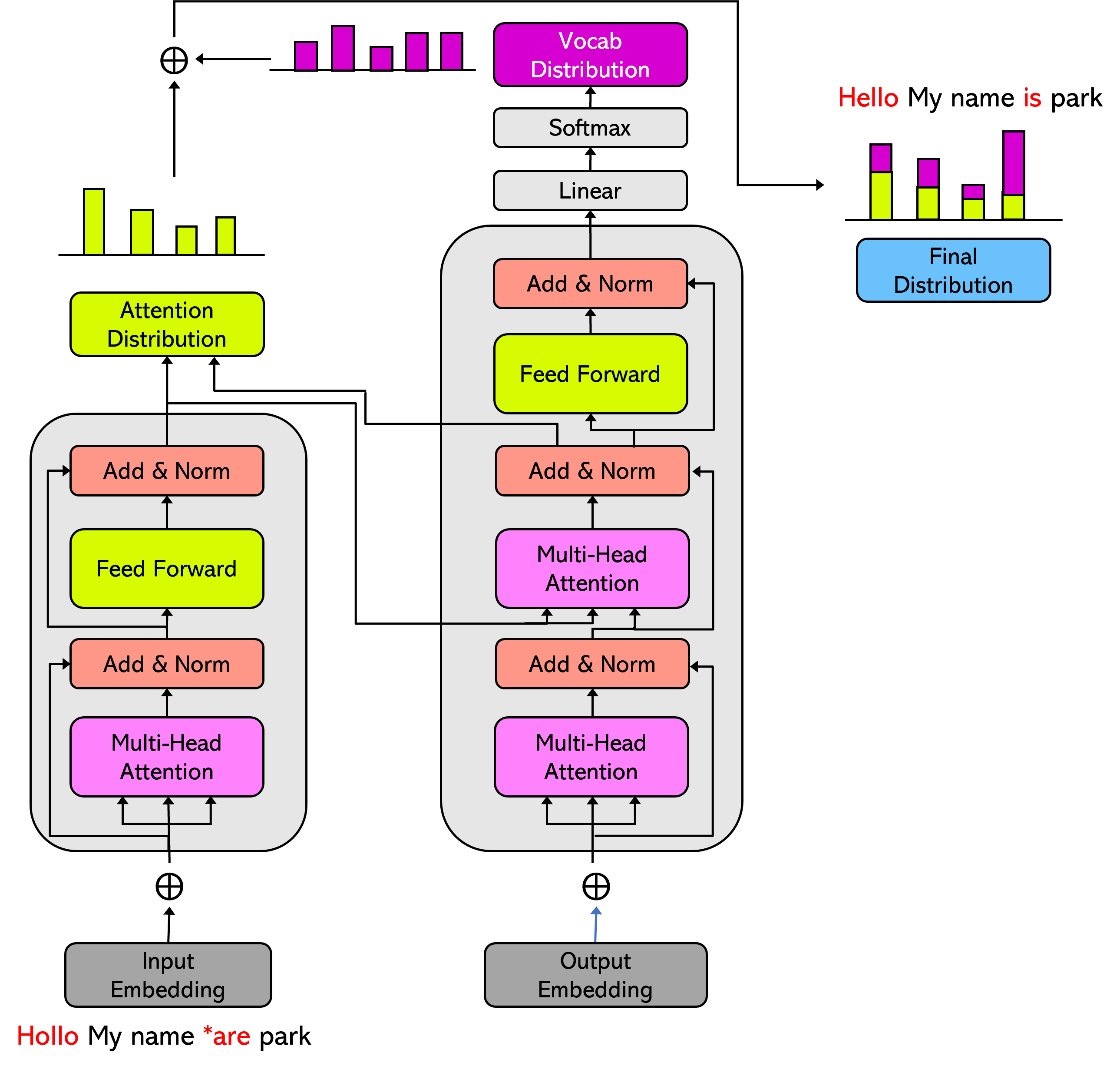}
\end{center}
\caption{Overview of grammar error correction model in \textsc{FreeTalky}.} 
\label{fig:GEC}
\end{figure}

\paragraph{Grammar Error Feedback} 
Several studies have proven the positive effects of feedback on error correction in foreign language learning~\cite{lyster1997corrective,loewen2013role,nassaji2016anniversary, park2021bts}. Inspired by these results, this study adapted the GEC model for \textsc{FreeTalky} to provide feedback on  grammatical errors to users. 

In terms of machine translation, GEC is a task in which the erroneous source sentence is translated to the correct target sentence. That is, the sequence-to-sequence (S2S) model can be interpreted as noise sequence to denoise sequence (NS2DS~\cite{park2020neural, park2020comparison}). All S2S models, such as CNN~\cite{gehring2017convolutional}, RNN~\cite{bahdanau2014neural}, and Transformer~\cite{vaswani2017attention}, can be utilized as NS2DS. In addition, existing rule-based and statistical models can be incorporated with NS2DS models in a hybrid form. 

In this study, we adopted the vanilla Transformer~\cite{vaswani2017attention} with a copy mechanism~\cite{gu2016incorporating}, which was suitable for the service owing to its lightweight structure. We applied the copy mechanism based on the characteristics of the training data in the GEC, where the source and target sentences had the same character set. Thus, we handled the out-of-vocabulary (OOV) problem using the copy mechanism in a principled manner. 

The overall model structure is shown in Figure~\ref{fig:GEC}. 
It was deployed on a separate server and connected to \textsc{FreeTalky} by means of REST API using Flask. To enable scaling-up of the service, we additionally deployed the CPU-based model, as well as the GPU-based model~\cite{klein2020efficient}. Through this function, explicit instructions are made available to the user. 

\begin{table*}[ht]
\centering
\resizebox{\textwidth}{!}{
\begin{tabular}{llc}
\toprule
\multicolumn{2}{c}{\textbf{Example conversation with \textsc{FreeTalky}}} \\
\midrule
FreeTalky: & Hi, I am NAO. Please choose the number between 1 to 3 and tell me, then I will set my personality & \texttt{(01)} \\ 
& and tell you about my persona! & \texttt{(02)}\\ 
User: & Two & \texttt{(03)}\\
FreeTalky: & Persona setting is done. Let me introduce myself. I live in colorado. I like to go hiking in the spring. & \texttt{(04)} \\
& My favorite activity is rock climbing. I am a mechanical engineer. & \texttt{(05)}\\
FreeTalky: & Now you can have a conversation with me! If you want to finish the conversation, just say bye. & \texttt{(06)}\\ 
User: & Hi how are you? & \texttt{(07)}\\
FreeTalky: & I am doing great! I just got back from rock climbing. What are you up to? & \texttt{(08)}\\
User: & Wow... That’s awesome. I am going to study English. & \texttt{(09)}\\
FreeTalky: & I am in colorado, we are going to go hiking sometime soon. & \texttt{(10)}\\
User: & \colorbox{pink}{I want visit Colorado.} I think you like outdoor activities! & \texttt{(11)}\\ 
FreeTalky: & I do, but I'm not too big on them. & \texttt{(12)}\\ 
User: & I see. \colorbox{lime}{You.. engineer?} & \texttt{(13)}\\
FreeTalky: & Yes, I am a mechanical engineer. & \texttt{(14)}\\
User: & Okay.. It’s nice to meet you. Bye. & \texttt{(15)}\\
FreeTalky: & Conversation is done. Do you want to get grammatical feedback? & \texttt{(16)}\\
User: & Yes please. & \texttt{(17)}\\
FreeTalky: & Okay! You said, \colorbox{pink}{“I want visit Colorado”}, but \colorbox{pink}{“I want to visit Colorado”}, is a more grammatically & \texttt{(18)}\\ 
& correct expression. You said, \colorbox{lime}{“You engineer?”}, but \colorbox{lime}{“Are you an engineer?”} is a more & \texttt{(19)}\\ 
& grammatically correct expression. & \texttt{(20)}\\
\bottomrule
\end{tabular}
}
\caption{An example conversation between the user and FreeTalky system. Users can set persona by choosing number and talk with FreeTalky system. It gives grammar error feedback to the user after chatting. Full video is attached in \textit{Supp.}. The highlighted text indicates the user's grammar error and FreeTalky's feedback sentence.}
\label{tab:conversation}
\end{table*}

\section{\textsc{FreeTalky} for Xenoglossophobia}
Xenoglossophobia, also known as foreign language anxiety, is a consistent and irrational fear revolving around foreign languages~\cite{abinaya2016xenoglossophobia}. People who suffer from xenoglossophobia experience anxiety while having conversations with people who speak a foreign language. There does not exist a specific method for treating xenoglossophobia in particular; however, the treatments suggested for general phobias can be utilized.

Most people who have a phobia are treated with exposure therapy. It is a method that involves exposing the patients to the situation, thus restricting the anxiety for a certain period. Though anxiety can temporarily increase during the treatment, it helps to release it in the long term, when the patients face real-world situations. 

Motivated by this treatment, we experimented to determine whether \textsc{FreeTalky} could practically help people who have xenoglossophobia. The NAO robot replaces the second human in the conversation, thus possibly relieving the anxiety of the user, and simultaneously makes the situation more realistic using the persona-based free talking system.

\section{Qualitative Analysis}
We analyzed the conversation between the user and \textsc{FreeTalky} in a qualitative manner. An example conversation is presented in Table \ref{tab:conversation}. As shown in line \texttt{(01)}, the FreeTalky system provides a randomly chosen persona through the selection of a number between 1 to 3. This setting enables virtual experiences in various situations where the users can experience the feeling of talking to different people. In lines \texttt{(08)} and \texttt{(14)}, \textsc{FreeTalky} generates persona-based answers. Its multi-turn conversation not only seems natural, but also continues until the user ends the conversation themselves. Moreover, it has the ability to ask an additional question from the user, as in \texttt{(08)}. Once the conversation is over, it provides grammatical error feedback to the user, as in the last utterance of \textsc{FreeTalky} in Table \ref{tab:conversation}. 
  
\begin{figure}
\begin{center}
\includegraphics[width=1.0\linewidth]{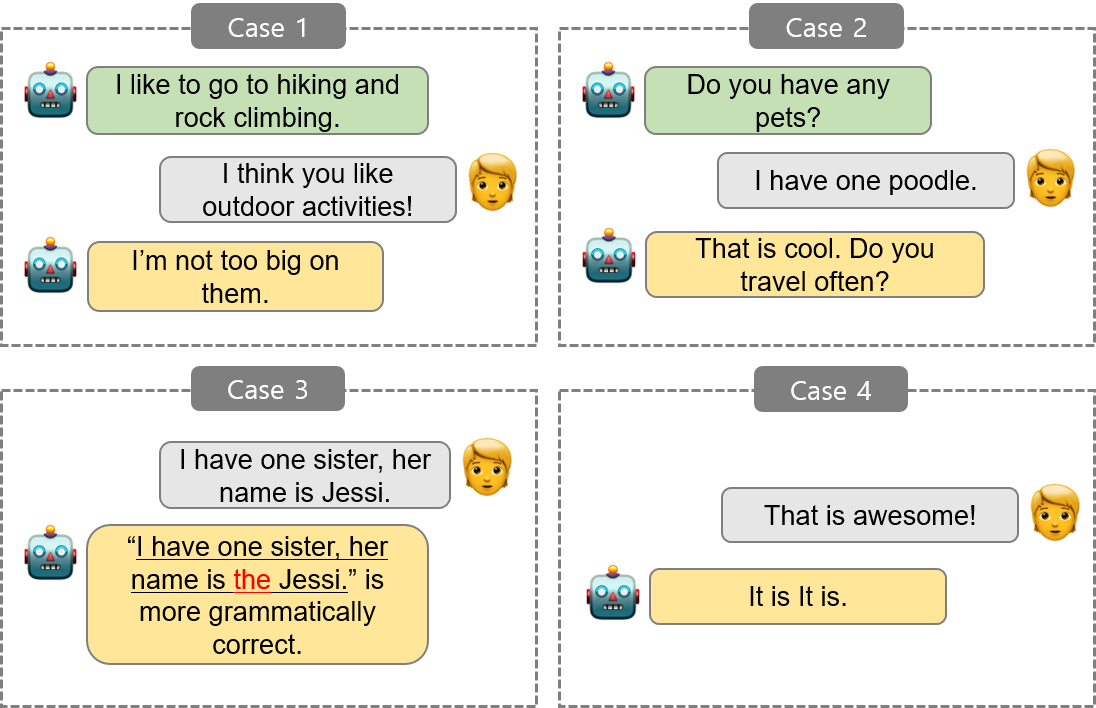}
\end{center}
\caption{Four cases of errors. Case 1 is an erroneous sentence that conflicts with what the model said before. Case 2 is an unnatural utterance that suddenly changes the conversation topic. Case 3 is an error of overcorrection. Case 4 is a generic problem of generative models. } 
\label{fig:error}
\end{figure}

\paragraph{Error Analysis}
Although the proposed system can hold a natural multi-turn dialogue and provide grammatical error feedback, several errors remain in need of improvement besides the given conversation. We divided the error types into four cases, as shown in Figure \ref{fig:error}. The first case is the error of unnatural flow of conversation. Although the model attempts to converse about specific topics, it says things that conflict with what was said before, or with its persona, as shown in line \texttt{(12)} of Table \ref{fig:freetalky}. The second case is the utterance pattern learned from the pre-trained model. As we utilized the pre-trained model, the pattern learned from the large dialogue corpus tends to appear, as a result of which, the model sometimes changes the conversation topic and talks about itself. In Figure \ref{fig:error}, the FreeTalky model suddenly changes the topic of conversation to travel. The third case is overcorrection. Overcorrection is a phenomenon wherein a grammatically correct input sentence is regarded as a subject that requires correction, and the sentence structure is disturbed. In Figure \ref{fig:error}, the model changes a correct sentence of the user. The fourth case is a generic problem of generative models, that is, a model tends to generate tokens repeatedly and leads to a \texttt{[unk]} token issue, as shown in Figure \ref{fig:error}.

\section{\textsc{FreeTalky} Satisfaction Survey} 
We validated the efficiency of the proposed system through five middle school students, filtered by measurement~\cite{al2014revisiting}, who experience foreign language anxiety. Further, we used six different questionnaires for system validation based on the Likert scale~\cite{likert1932technique} that rates situations from 1 ({\em{i.e.,}} strongly disagree) to 5 ({\em{i.e.,}} strongly agree), as shown in Table~\ref{tab:questionnaire}. 

\begin{figure}[ht!]
\begin{center}
\includegraphics[width=0.85\linewidth]{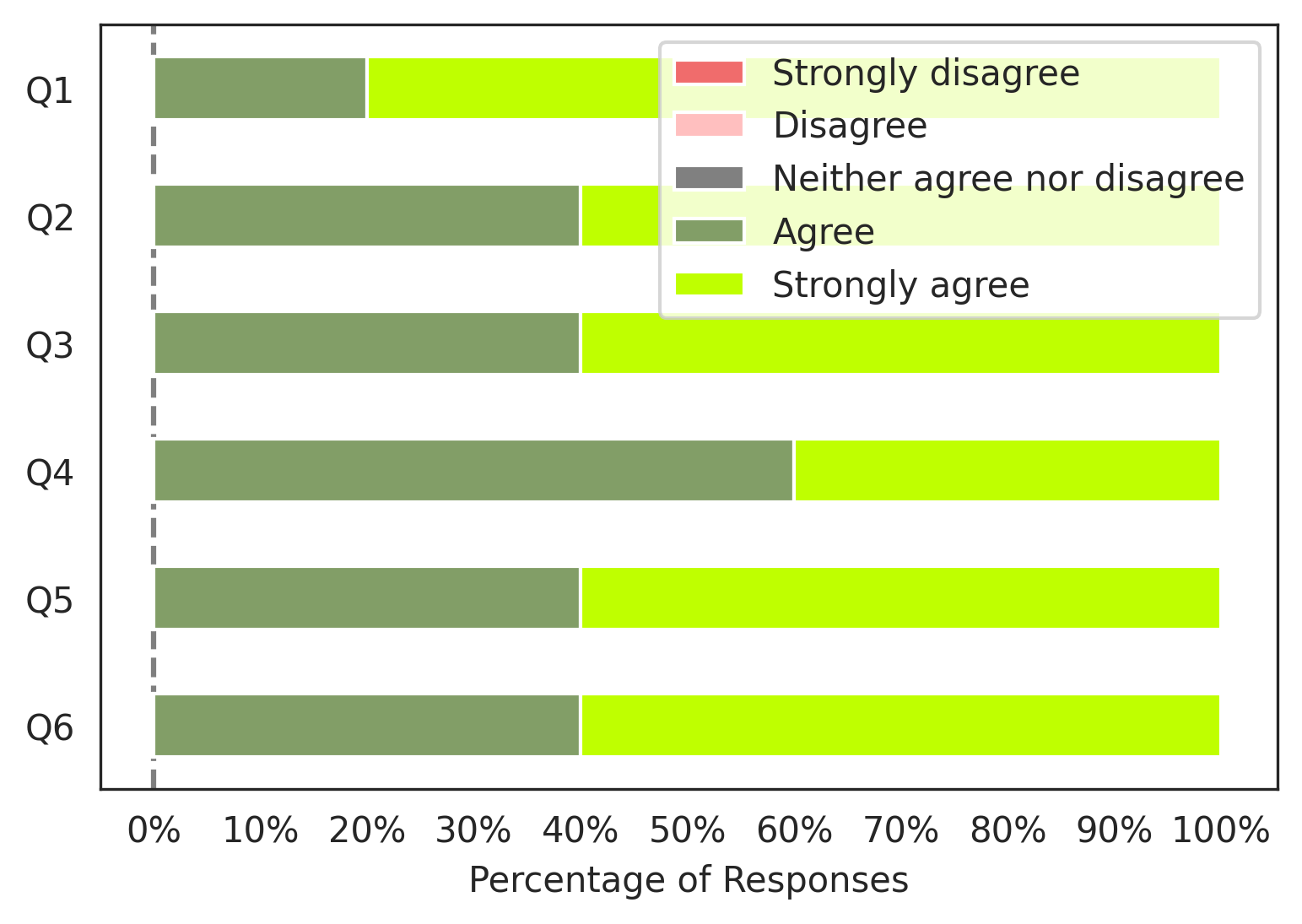}
\end{center}
\caption{Response result of satisfaction survey of \textsc{FreeTalky}.} 
\label{fig:satisfaction_response}
\end{figure}

As shown in Figure~\ref{fig:satisfaction_response}, most questions about the system were scored as strongly agree or agree. This result implies that the proposed system is suitable for alleviating people’s xenoglossophobia.

\renewcommand{\arraystretch}{1.2}
\begin{table}[tbh]
\centering
\resizebox{\columnwidth}{!}{
\begin{tabular}{ll}
\toprule
\multicolumn{2}{c}{\textbf{Questionnaire}}\\ \hline \hline
Q1 &  I feel comfortable while interacting with a robot in \\ 
&  a real human-like conversation. \\ \hline
Q2 & I do not feel scared while talking to a robot. \\ \hline
Q3 & Do you think \textsc{FreeTalky} will help alleviate your \\ 
&  foreign language anxiety? \\ \hline
Q4 & Are you satisfied with the performance of grammatical \\ 
& error feedback? \\ \hline
Q5 & Do you think \textsc{FreeTalky} performs naturally in a \\
& multi-turn conversation? \\ \hline
Q6 & Do you think persona is well applied in \textsc{FreeTalky}? \\ 
\bottomrule
\end{tabular}}
\caption{\label{tab:questionnaire} Questionnaire of \textsc{FreeTalky} satisfaction survey.}
\end{table}

\begin{table*}[h!]
\centering
\resizebox{\textwidth}{!}{
\begin{tabular}{l|l}
\toprule
\multicolumn{2}{c}{\textbf{User interview before chatting with \textsc{FreeTalky}.}} \\
\midrule
Interviewer & Hi, nice to meet you. \\
& I heard that you experience fear while meeting and conversing with foreigners. \\
User & That is right. I have learned English, but when I meet foreigners, I feel nervous, so I am unable to talk to them. \\ 
Interviewer & Okay, do you face any problems while reading and writing in English? \\
User & Yes, reading text in English is not a problem, but I get very nervous when I speak in English. \\
Interviewer & Have you made any efforts to overcome this? \\ 
User & I intended to go to an English conversation academy, but the cost was expensive, and I do not have enough \\ 
& time to go there. \\
Interviewer & Okay, then how about practicing conversing in English with a robot, instead of a human? \\
User & A robot? I have not talked to a robot before, so I do not know. \\
Interviewer & Then, why don’t we try and see if it can help relieve your fear now? \\
\midrule
\multicolumn{2}{c}{\textbf{User interview after chatting with \textsc{FreeTalky}.}} \\ 
\midrule
Interviewer & How was it, talking with the robot in English?  \\
User & It was good because I felt more relaxed than when I talk to a real person. Also, it was good that this \\
& robot gave me feedback on grammatically wrong expressions. \colorbox{yellow}{The robot had various voices, so I felt like} \\
& \colorbox{yellow}{it was a real situation.} \\
Interviewer & You mean that talking with the robot helps relieve your psychological fear. \\
User & Yes, it does. \\
Interviewer & Great then, we will attempt to further improve this system to assist with your anxiety. Thank you \\
User & Thank you. \\
\bottomrule
\end{tabular}}
\caption{User interview and experience of chatting with \textsc{FreeTalky}. As the user interview was conducted in Korean, the translated version is attached. The highlighted text indicates the notable points in the interview (see details in discussion).}
\label{tab:interview}
\end{table*}

\begin{figure}
\begin{center}
\includegraphics[width=0.85\linewidth]{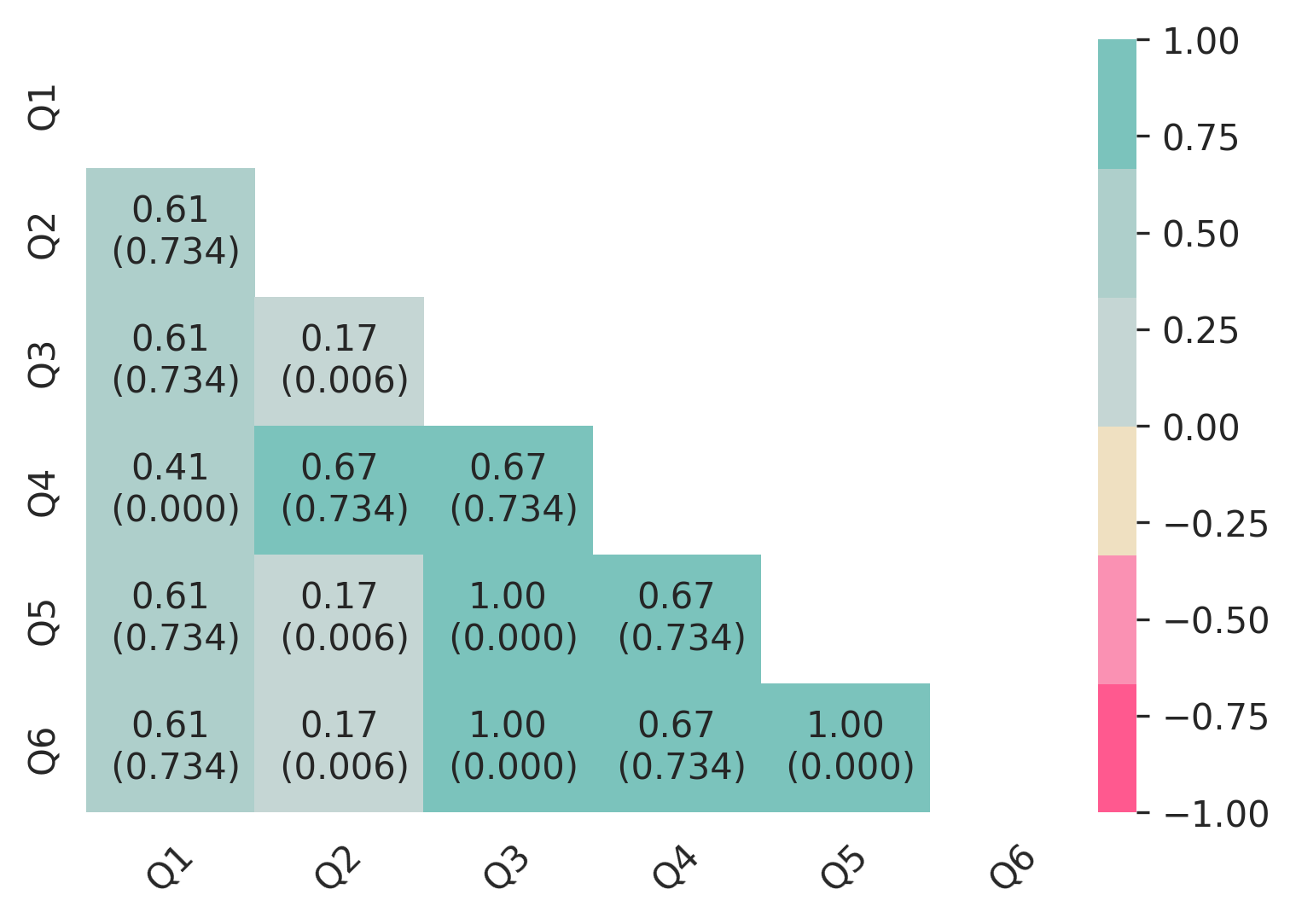}
\end{center}
\caption{Correlation result of satisfaction survey of \textsc{FreeTalky}. Each cell indicates a correlation value, and numbers in parentheses are p-value (p-value$<$0.05 indicates statistical significance). } 
\label{fig:satisfaction}
\end{figure}

Furthermore, we conducted Spearman correlation~\cite{de2016comparing} on the survey results in Figure~\ref{fig:satisfaction}. The results show a positive correlation in all question relationships. There is a significant (p$<$0.05) positive correlation between Q1 and Q4, from Q2 to Q3, Q5, Q6, and between Q5 and Q6. Interestingly, Q3 and Q5, Q6 show a perfect correlation, which implies that there is a high association between the alleviation of foreign language anxiety and natural multi-turn dialogue and persona-applied conversation. These results prove that \textsc{FreeTalky} is appropriate for our objectives, that applying the persona and multi-turn dialogue to the conversation would result in the system being more user-friendly. Another highlight of the results corresponding to our objectives is that the humanoid robot NAO performed well in alleviating xenoglossophobia, and is user-friendly according to Q2, Q3, Q5, and Q6. Additionally, it demonstrated that GEC contributes to the comfort of the users while interacting with humanoid robots, according to the results of Q1 and Q4. Consequently, these factors determine the system satisfaction.

\section{Discussion}
Following the system satisfaction survey, we interviewed the actual users to determine an in-depth usability of the system besides the survey. 
The interview process aided the system development, as a predefined questionnaire could not handle unexpected answers, such as a narrative context. The interview transcript is presented in Table~\ref{tab:interview}. The original interview video is also attached in the following link\footnote{\url{https://youtu.be/P6Xyv3pi6F4}}. 

As illustrated in Table~\ref{tab:interview}, we determined that the user experienced a life-like situation owing to the various voices of NAO (see yellow box in Table~\ref{tab:interview}). We provided two modulations of voices for NAO, that is \textit{neutral} and \textit{joyful}. The user in this interview experienced NAO as a \textit{joyful} voice; thus, this setting enabled the user to experience a realistic situation. 

Although we performed the fundamental quantitative analysis and appropriate qualitative analysis, limitations that hindered the quantitative analysis of the system remain because there is no official test set available for further evaluation. Thus, we focused on the error analysis to address these limitations. However, these limitations encourage our further research, that is, to build a test set for evaluating the system and develop an evaluation metric. We plan to extend our work and release the same, to support researchers in this field.    

\section{Conclusion}
We proposed \textsc{FreeTalky}, where users can receive feedback on English grammar errors and improve their English conversation skills. In addition, as \textsc{FreeTalky} is based on a persona dialogue system, the content of multi-turn conversations can be adaptively altered to suit the users and provide diverse conversational experience to the users. Furthermore, it is possible to alleviate the psychological pressure experienced by people suffering from xenoglossophobia, using the NAO robot. 
%In the future, we intend to provide a customized educational conversation system service in the foreign language learning business, using data for various language pairs other than only English. 

\section*{Acknowledgment}
This research was supported by the MSIT(Ministry of Science and ICT), Korea, under the ITRC(Information Technology Research Center) support program(IITP-2018-0-01405) supervised by the IITP(Institute for Information \& Communications Technology Planning \& Evaluation) and supported by the MSIT(Ministry of Science and ICT), Korea, under the ICT Creative Consilience program(IITP-2021-2020-0-01819) supervised by the IITP(Institute for Information \& communications Technology Planning \& Evaluation).

\bibliography{aaai22}

\end{document}